\documentclass[12pt,longbibliography,eprint]{revtex4-1}
\usepackage{epigraph}
\usepackage{graphicx}
\usepackage{epstopdf}
\usepackage{hyperref}
\usepackage{numprint}

\begin{document}
\preprint{V.M.}
\title{Norm--Free Radon--Nikodym Approach to Machine Learning}
\author{Vladislav Gennadievich \surname{Malyshkin}} 
\email{malyshki@ton.ioffe.ru}
\affiliation{Ioffe Institute, Politekhnicheskaya 26, St Petersburg, 194021, Russia}

\date{December, 10, 2015}

\begin{abstract}
\begin{verbatim}
$Id: MLVector.tex,v 1.85 2015/12/15 19:00:05 mal Exp $
\end{verbatim}
For Machine Learning (ML) classification problem,
where a vector of $\mathbf{x}$--observations (values of attributes)
is mapped to a single $y$ value (class label),
a generalized Radon--Nikodym type of solution is proposed.
Quantum--mechanics --like probability states $\psi^2(\mathbf{x})$ are
considered and ``Cluster Centers'', corresponding to the
extremums of $<y\psi^2(\mathbf{x})>/<\psi^2(\mathbf{x})>$,
are found from generalized eigenvalues problem.
The eigenvalues give possible $y^{[i]}$ outcomes and
corresponding to them eigenvectors $\psi^{[i]}(\mathbf{x})$ define ``Cluster Centers''.
The projection of a $\psi$ state, localized at given $\mathbf{x}$ to classify,
on these
eigenvectors define the probability
of $y^{[i]}$ outcome, thus avoiding using a norm ($L^2$ or other types),
required for ``quality criteria'' in
a typical Machine Learning technique.
A coverage of each `Cluster Center''
is calculated, what potentially allows to separate
system properties (described by $y^{[i]}$ outcomes)
and system testing conditions (described by $C^{[i]}$ coverage).
As an example of such application $y$ distribution estimator
is proposed in a form of pairs $(y^{[i]},C^{[i]})$, that
can be considered as Gauss quadratures generalization.
This estimator allows to perform $y$ probability distribution estimation
in a strongly non--Gaussian case.
\end{abstract}

\maketitle

\section{\label{intro}Introduction}
Machine Learning(ML) explores the study and construction of algorithms that can learn from and make predictions on data.
The key four elements\cite{maloldarxiv} of any ML model
is (1) Attribute selection. (2) Knowledge Representation.
(3) Quality Criteria. (4) Search algorithm.
The first three elements are the most important in practice,
but search algorithms often attract most attention of ML researchers.
In this work we will try to address
the fist three elements.
The main idea of this work is to find observations ``Cluster Centers''
as corresponding to matrix spectrum of class label,
and then project the state to classify
on these ``Cluster Centers'', thus receive probabilities directly
and avoid using a norm for quality criteria.

\section{\label{QSol}Generalized Radon--Nikodym Solution}

Consider the following ML problem
where attributes vector of $d_x$ components
is mapped to a single outcome (class label) observation $y$ for $l=[1..M]$.
\begin{eqnarray}
  (x_0,x_1,\dots,x_m,\dots,x_{d_x-1})^{(l)}&\to&y^{(l)}  \label{mlproblem}
\end{eqnarray}
A number of other problems can be converted to this problem,
e.g. distribution regression problem\cite{2015arXiv151109058G},
can be converted by using bag's distribution moments as $x_m$ vector components.
A lot of ML theories are of interpolatory type,
where the (\ref{mlproblem}) is piecewise interpolated
by regression coefficients, propositional rules, decision trees
or Neural Networks minimizing some norm--like quality criteria,
see \cite{witten2005data} for excellent review and implemented algorithms.
But we are going to treat the (\ref{mlproblem}) not in terms
of some error minimization, but 
probabilistically.
Consider the $\psi(\mathbf{x})$ state
\begin{eqnarray}
  \psi(\mathbf{x})&=&\sum\limits_{m=0}^{d_x-1}\psi_m x_m
  \label{psidef}
\end{eqnarray}
that is defined by $\psi_m ; m=[0..d_x-1]$ vector.
Any more complex forms of $\psi$  (e.g. some functions $f(x_m)$)
is equivalent
to adding $f(x_m)$ terms to $\mathbf{x}$ vector (\ref{mlproblem}),
and the form (\ref{psidef}) pose no limitation,
because any of such can be avoided by
adding more $x$-- components.

Introduce
the measure $\sum_{l=1}^{M}$ and treat the $\psi^2$
as ``probability density''. Consider corresponding $y_{\psi}$
\begin{eqnarray}
  y_{\psi}&=&\frac{\sum\limits_{l=1}^{M} y^{(l)}  \psi^2(\mathbf{x}^{(l)})}
  {\sum\limits_{l=1}^{M} \psi^2(\mathbf{x}^{(l)})}
  \label{ypsi} \\
  <f>&=& \sum\limits_{l=1}^{M} f^{(l)} \\
  \left( G \right)_{qr} &=& \left<x_q x_r\right>= \sum\limits_{l=1}^{M} x_q^{(l)} x_r^{(l)} \label{G}\\
  \left( yG \right)_{qr} &=& \left<y x_q x_r\right>= \sum\limits_{l=1}^{M} y^{(l)} x_q^{(l)} x_r^{(l)} \label{yG}\\  
  y_{\psi}&=& \frac{\sum\limits_{q,r=0}^{d_x-1} \psi_q \left( yG \right)_{qr} \psi_r}
  {\sum\limits_{q,r=0}^{d_x-1} \psi_q \left( G \right)_{qr} \psi_r}
\end{eqnarray}

The $\psi^{[i]}$ states, corresponding to the extremums of $y_{\psi}$
can be found from generalized eigenvectors problem
\begin{eqnarray}
  \sum\limits_{r=0}^{d_x-1}\left(yG\right)_{qr} \psi^{[i]}_r &=& y^{[i]} \sum\limits_{r=0}^{d_x-1}\left(G\right)_{qr} \psi^{[i]}_r
  \label{gevproblem}
\end{eqnarray}
The $y^{[i]}$ give possible outcomes and
\begin{eqnarray}
  \psi^{[i]}(\mathbf{x})&=&\sum\limits_{m=0}^{d_x-1}\psi^{[i]}_m x_m
  \label{psidefcl}
\end{eqnarray}
define ``Cluster Centers'', corresponding to $y^{[i]}$. The value
of $\left(\psi^{[i]}(\mathbf{x}^{(l)})\right)^2$ is typically large
only for the $l$'s at which $y^{(l)}$  value is close to the eigenvalue $y^{[i]}$.
Note that
\begin{eqnarray}
  &&\sum\limits_{q,r=0}^{d_x-1}\psi^{[j]}_r\left(yG\right)_{qr} \psi^{[i]}_r
  =y^{[i]} \delta_{ji} \\
  &&\sum\limits_{q,r=0}^{d_x-1}\psi^{[j]}_r\left(G\right)_{qr} \psi^{[i]}_r
  = 
  \sum\limits_{l=1}^{M}\psi^{[j]}(\mathbf{x}^{(l)}) \psi^{[i]}(\mathbf{x}^{(l)})=
  \delta_{ji}
  \label{probsum}
\end{eqnarray}
The (\ref{probsum}) allows to treat the
$\left(\psi^{[i]}(\mathbf{x}^{(l)})\right)^2$
as the value proportional to the probability of $l$--th
learning observation from (\ref{mlproblem})
to have the $y^{[i]}$ outcome.
Similarly for two given $\mathbf{x}^A$ and $\mathbf{x}^B$
their projection to each other
\begin{eqnarray}
<\mathbf{x}^A|\mathbf{x}^B>_{\pi}&=&\sum\limits_{q,r=0}^{d_x-1}
x_q^A\left(G\right)^{-1}_{qr} x_r^B
  \label{proj}
\end{eqnarray}
The probabilities, calculated by projecting the
given $\mathbf{x}$ to ``Cluster Centers'' 
are:
\begin{eqnarray}
  w^{[i]}(\mathbf{x})&=&\left(\sum_{r=0}^{d_x-1}x_r  \psi^{[i]}_r \right)^2
  \label{wi}
  \\
  P^{[i]}(\mathbf{x})&=&w^{[i]}(\mathbf{x})/\sum_{r=0}^{d_x-1} w^{[r]}(\mathbf{x}) \label{Pi}
\end{eqnarray}
This is the solution to classification problem:
for a given $\mathbf{x}$
the eigenvalues $y^{[i]}$ from (\ref{gevproblem})
provide possible outcomes and $P^{[i]}(\mathbf{x})$
from (\ref{Pi}) provide each outcome probability.
This answer is much more general than, say, regression type
of answer, in which only $y$ estimate can be given
and probability distribution can be estimated from
standard deviation only for Gaussian type of random variables.
The (\ref{gevproblem}) does not use second $y$ moment at all,
so the answer can be successfully applied
to non--Gaussian samples, e.g. the ones with infinite standard deviation
of $y$.

If $y$ estimate for a given $\mathbf{x}$ is required two answers can be provided,
see \cite{2015arXiv151005510G} Appendix D,
Least Squares $A_{LS}$ and Radon--Nikodym $A_{RN}$. The answers are:
\begin{eqnarray}
  Y_q&=&\sum_{l=1}^{M} y^{(l)} x_q^{(l)} \label{Yq} \\
  A_{LS}(\mathbf{x})&=& \sum\limits_{q,r=0}^{d_x-1} x_q \left(G\right)^{-1}_{qr} Y_r \label{ALS} \\
  A_{RN}(\mathbf{x})&=& \frac{\sum\limits_{q,r,s,t=0}^{d_x-1} x_q \left(G\right)^{-1}_{qr} \left(yG\right)_{rs}
    \left(G\right)^{-1}_{st}  x_t}
  {\sum\limits_{q,r=0}^{d_x-1}x_q \left(G\right)^{-1}_{qr} x_r} \label{ARN}
\end{eqnarray}
The (\ref{ALS}) is least squares answer to $y$ estimation given $\mathbf{x}$.
The (\ref{ARN}) is Radon--Nikodym answer to $y$ estimation given $\mathbf{x}$.
These answers can be considered as an extension of least squares and Radon--Nikodym type
of answers to vector input.
In case $x_m$ components in (\ref{mlproblem})
are the moments of some random variable the
$A_{LS}$ and $A_{RN}$ are reduced exactly to the problem
of learning from random distribution
we considered in Ref. \cite{2015arXiv151109058G}.
Note, that the $A_{LS}(x)$ answer not necessary
preserve $y$ sign, but  $A_{RN}(x)$ always preserve $y$ sign,
same as we have in our earlier works.

One more issue we want to discuss is coverage estimation for each ``Cluster Center''
$\psi^{[i]}(\mathbf{x})$.
The states $\psi^{(l)}(\mathbf{x})$, corresponding to specific $\mathbf{x}^{(l)}$
is normalized projection (\ref{proj}) with $\mathbf{x}=\mathbf{x^A}$ and $\mathbf{x}^{(l)}=\mathbf{x^B}$,
\begin{eqnarray}
  \psi^{(l)}(\mathbf{x})&=&  \frac{\sum\limits_{q=0}^{d_x-1}
    x_q \left(G\right)^{-1}_{qr} x^{(l)}_r}{
    \sqrt{\sum\limits_{q,r=0}^{d_x-1}x^{(l)}_q \left(G\right)^{-1}_{qr} x^{(l)}_r  }}
  \label{psil} \\
  \omega^{(l)}_{[i;j]}&=&
  \frac{\sum\limits_{q=0}^{d_x-1} \psi^{[i]}_q x^{(l)}_q
    \sum\limits_{q=0}^{d_x-1} \psi^{[j]}_q x^{(l)}_q }
       {\sum\limits_{q,r=0}^{d_x-1} x^{(l)}_q  \left(G\right)^{-1}_{qr} x^{(l)}_r}
       \nonumber \\
       &=&<\psi^{[i]}(\mathbf{x})|\psi^{(l)}(\mathbf{x})><\psi^{(l)}(\mathbf{x})|\psi^{[j]}(\mathbf{x})>
       \label{omegaw} \\ 
       C^{[i]}&=&\sum\limits_{l=1}^{M}\omega^{(l)}_{[i;i]}  \label{Ccoverage} \\
       D^{[i]}&=&\sum\limits_{l=1}^{M}\omega^{(l)}_{[i;i]}\left(1-\omega^{(l)}_{[i;i]}\right)
       =\sum\limits_{l=1}^{M}\sum\limits_{j=0;j\ne i}^{d_x-1}
       \left(\omega^{(l)}_{[i;j]} \right)^2
       \label{wdisclass}
\end{eqnarray}
can be projected  to each $\psi^{[i]}(\mathbf{x})$
and the square of the projection give the probability $\omega^{(l)}_{[i;i]}$
of $l$--th training observation to have the value of outcome equal to $y^{[i]}$.
The sum over
all training $l=[1..M]$ observations give $i$--th
Cluster Center coverage $C^{[i]}$, the ``effective number'' of training
observations covered by the $i$--th cluster, Eq. (\ref{Ccoverage}).
Note that $\sum\limits_{i=0}^{d_x-1}\omega^{(l)}_{[i;i]}=1$ and 
$\sum\limits_{i=0}^{d_x-1}C^{[i]}=M$.
The $D^{[i]}$ from (\ref{wdisclass}), cluster localization measure,
determine how ``pure'' the $i$--th cluster center $\psi^{[i]}(\mathbf{x})$ is. For a fixed $l$
each term in (\ref{wdisclass})
is a product of  probabilities that $l$--th observation is in the
$i$--th state $\omega^{(l)}_{[i;i]}$ and also is
in a $j\ne i$ state $\sum\limits_{j=0;j\ne i}^{d_x-1}\omega^{(l)}_{[j;j]}=1-\omega^{(l)}_{[i;i]}$.
Cluster relative localization (lower than 1 value) can be also considered as $D^{[i]}/C^{[i]}$.

This  approach produce $d_x$ clusters, and for each $i=[0..d_x-1]$
eigenvalue $y^{[i]}$,
coverage $C^{[i]}$ and
localization measure $D^{[i]}$ are obtained.
The question about selection of ``true'' clusters arise.
Among the $x_m; m=[0..d_x-1]$ components of (\ref{mlproblem})
many may be irrelevant (will be weeded out, not a problem)
or redundant (lead to $G_{qr}$ matrix (\ref{G}) degeneracy
and require regularization of the problem).
A criteria can be applied,
that select only some best $d\le d_x$ components as linear
combination of $d_x$ components, initially available in (\ref{mlproblem}).
The power of our approach is that the
generalized eigenvalues equation (\ref{gevproblem})
can be used not only for possible $y$ outcomes estimation,
but also for components selection
using various criteria.
As illustration let us find the $d\le d_x$ components,
providing maximal coverage $C$.
Consider a (\ref{psidef}) state, then using the (\ref{Ccoverage})
definition, the coverage $C_{\psi}$, corresponding to a $\psi(\mathbf{x})$ state,
can be calculated as
\begin{eqnarray}
  n^{(l)}&=&\frac{1}{\sum\limits_{k,m=0}^{d_x-1} x^{(l)}_k  \left(G\right)^{-1}_{km} x^{(l)}_m}  \label{nl} \\  
  \left( CG \right)_{qr} &=& \sum\limits_{l=1}^{M}  n^{(l)} x_q^{(l)} x_r^{(l)}
   \label{CG}\\  
  C_{\psi}&=&\frac{\sum\limits_{q,r=0}^{d_x-1} \psi_q \left( CG \right)_{qr} \psi_r}
  {\sum\limits_{q,r=0}^{d_x-1} \psi_q \left( G \right)_{qr} \psi_r}
  \label{Cpsi} \\
  \sum\limits_{r=0}^{d_x-1}\left(CG\right)_{qr} \psi^{[j]}_r &=& C^{[j]} \sum\limits_{r=0}^{d_x-1}\left(G\right)_{qr} \psi^{[j]}_r
  \label{Cgevproblem} \\
  M&=&\sum\limits_{j=0}^{j=d_x-1}C^{[j]}
\end{eqnarray}
The Eq. (\ref{Cgevproblem}) is exactly (\ref{gevproblem})
but, instead of matrix $\left(yG\right)_{qr}$,
the matrix $\left( CG \right)_{qr}$ is used,
and eigenvectors $\psi^{[j]}_r$ from (\ref{Cgevproblem})
correspond to the states having coverage extremums.
Selecting $d\le d_x$ $\psi^{[j]}_r$ states,
corresponding to maximal $C^{[j]}$ would give
the required $d$ states with maximal coverage.

Similar to coverage, a number of observations
falling within given $y$ interval,
can be obtained by modification of (\ref{nl}):
take the $l$--th observation only when $y^{(l)}$
fall within the given $y$ interval.
This way an entropy of $y$ distribution
for a given $\psi(\mathbf{x})$ state,
can be easily obtained. However, the expression
for the entropy is not a plain ratio
of two quadratic form on $\psi(\mathbf{x})$
components, so the problem
(\ref{gevproblem})
cannot be directly applied to entropy calculation in general case.
But in the case of a discrete $y^{(l)}$, taking only two class values,
a classifier, maximizing 
the difference in outcomes number between the two
classes can be readily obtained.
Modify the (\ref{nl}) by taking
the terms with $y^{(l)}$ in class 1 using positive sign
and with $y^{(l)}$ in class 2 using negative sign.
Then the (\ref{Cpsi}) provide the difference
in observations number between class 1 and class 2 
as a ratio of two quadratic forms  on $\psi(\mathbf{x})$ components.
Solve (\ref{Cgevproblem}),
the eigenvalues now the strength of class prediction
and eigenvectors provide the classifiers.
Usage,
instead of $y^{(l)}$, that take two class values,
the difference in observations number between class 1 and class 2
allows to overcome $y$-- spectrum degeneracy, arising
from only two $y^{(l)}$ outcomes in a discrete case of
the problem.

One more important advantage of
using generalized eigenvalues problem (\ref{gevproblem})
is that the solution is
stable with respect to $y^{(l)}$ outlier observations.
Such observations give, for some limited number of $i$ values,
``Cluster Centers'' with outlier
eigenvalue
$y^{[i]}$
and small  coverage $C^{[i]}$.
These ``Cluster Centers''
with small coverage
can be either treated as outliers and
disregarded in case of measurement
errors presence,
or, typically more reasonable approach,
used for fat tails estimation of $y$ probability distribution.
Numerical experiments show,
that few $y^{(l)}$ outliers typically skew
spectrum and coverage
only for a single $i$ value,
leaving the rest of $(y^{[i]},C^{[i]})$ pairs intact. This is
drastically different from $y$--norm based approaches,
where a single $y^{(l)}$ outlier can easily skew a $L^2$ norm,
like standard deviation.

\subsection{\label{QMA}Quantum Mechanics Analogy}
 Quantum Mechanics interpretation can be given to this ML approach.
Consider some time--independent quantum system with
 Hamiltonian $H$, described by a state with $d_x$ components
 $\mathbf{x}=\{x_m\}; m=[0..d_x-1]$ of unknown nature.
Assume they are unknown, but stable
combinations of coordinate, momentum and angular momentum
and quantum system occupy a quantum
state, that change from one classical measurement to another.
The problem: From a number of measurement experiments
recover information about quantum system and experimental conditions.

Every classical measurement for such system
is a set of pairs (state,energy): $(\mathbf{x}^{(l)},E^{(l)})$,
index $l=[1..M]$ label classical measurement experiment.
It is clear that any kind of piecewise interpolation of $E(\mathbf{x})$
make no sense, because of possible degeneracy
of parameters $\mathbf{x}$ and
having a different mixed quantum state
in each measurement experiment. However, the $E^{(l)}$
correspond to quantum system Hamiltonian $H$.
The idea is: given large enough observations number $M$
select the experiments that are
close to ``pure quantum state''.
Introduce a wavefunction
$\psi(\mathbf{x})=\sum\limits_{m=0}^{d_x-1}\psi_m x_m$.
The entire observation set $l=[1..M]$ can be considered as a
quantum mechanics mixed state described by 
the density matrix
\begin{equation}
\rho(\mathbf{x^A},\mathbf{x^B})=
\sum\limits_{q,r=0}^{d_x-1}x_q^A \left( G \right)_{qr}^{-1} x_r^B
\label{densmatr}
\end{equation}
normalized for convenience on the number of components
$\left<\rho(\mathbf{x},\mathbf{x})\right>=Tr(\rho)=\sum\limits_{q,r=0}^{d_x-1} \left( G \right)_{qr} \left( G \right)_{rq}^{-1}=d_x$,
not to 1, like regular quantum mechanics density matrix.
Note, that
convenient in applications matrix averages, see Ref. \cite{2015arXiv151005510G} Appendix E,
are the averages, calculated on mixed state
with the density matrix (\ref{densmatr}), e.g.
\begin{equation}
  y_{\rho}=\sum\limits_{q,r=0}^{d_x-1}\left( yG \right)_{qr} \left( G \right)_{rq}^{-1}/d_x =\frac{\left<y\rho(\mathbf{x},\mathbf{x})\right>}
  {\left<\rho(\mathbf{x},\mathbf{x})\right>}
  \label{yrho}
\end{equation}
Reproducing Kernel (\ref{proj}) is plain density matrix
 $\rho(\mathbf{x^A},\mathbf{x^B})=<\mathbf{x}^A|\mathbf{x}^B>_{\pi}$.
The energy, corresponding to a pure state $\psi(\mathbf{x})$,
is $E_{\psi}=\sum\limits_{l=1}^{M} E^{(l)}  \psi^2(\mathbf{x}^{(l)})\Big/\sum\limits_{l=1}^{M}  \psi^2(\mathbf{x}^{(l)})$.
With a replacement of $E$ by $y$ we receive exactly the
problem (\ref{ypsi}).
Were we only know the $y_{\rho}$ from (\ref{yrho}), corresponding to the mixed state (\ref{densmatr}),
no information about system pure states can be obtained.
However, the problem (\ref{mlproblem}) have the $y^{(l)}$ outcome available
for each measurement experiment and 
generalized eigenvalues
problem (\ref{gevproblem}) now allows to estimate Hamiltonian spectrum,
then a projection (\ref{piproj}) of a given state $\psi^{(c)}(\mathbf{x})$
to eigenvectors of (\ref{gevproblem})
allows to estimate  $i$--th Hamiltonian state
contribution:
\begin{eqnarray}
  \psi^{[i]}(\mathbf{x})&=&\sum\limits_{q=0}^{d_x-1} x_q \psi^{[i]}_q \\
  \psi^{(c)}(\mathbf{x})&=&  \frac{\sum\limits_{q=0}^{d_x-1}
    x_q \left(G\right)^{-1}_{qr} x^{(c)}_r}{
    \sqrt{\sum\limits_{q,r=0}^{d_x-1}x^{(c)}_q \left(G\right)^{-1}_{qr} x^{(c)}_r  }}
\label{psiC}
  \\  
  <\psi^{(c)}|\psi^{[i]}>&=&
  \frac{
    \sum\limits_{q=0}^{d_x-1} x^{(c)}_q\psi^{[i]}_q}
       {\sqrt{\sum\limits_{q,r=0}^{d_x-1}x^{(c)}_q \left(G\right)^{-1}_{qr} x^{(c)}_r }}
\label{piproj}
       \\
  \psi^{(c)}(\mathbf{x})&=&
  \sum\limits_{q=0}^{d_x-1} <\psi^{(c)}|\psi^{[i]}> \psi^{[i]}(\mathbf{x})
  \label{psiexp}
\end{eqnarray}
If one put $x^{(c)}_m=x^{(l)}_m$ to (\ref{psiC}) then
(\ref{psiexp}) give $l$--th experiment wavefunction expanded
over the Hamiltonian states $\psi^{[i]}(\mathbf{x})$
and the $\left(<\psi^{(l)}|\psi^{[i]}>\right)^2=\omega^{(l)}_{[i;i]}$ give $i$--th outcome probability
for $l$--th experiment.
The Hamiltonian spectrum $y^{[i]}$ eigenvalues are
the characteristics of quantum system itself,
but the coverage  (\ref{Ccoverage})
provide information how often the $\psi^{[i]}$ pure
state give substantial contribution to 
mixed quantum state (\ref{densmatr}) of classical experiment, i.e.
about experiment conditions, not about quantum system itself.

The simplest ML application of this approach
can be just  to put into (\ref{psiC})
as a $x^{(c)}_m$ not the state $x^{(l)}_m$, corresponding to $l$--th training
datapoint, but the state we want to classify
(this state is exactly (\ref{psiC}),
with $x^{(c)}_m$ components equal to 
the components of $\mathbf{x}$ vector we want to classify)
in a hope to receive
a reasonable prediction.

But such a direct approach,
very much typical for ML applications, looks more like
of interpolatory type. However, there is a much deeper
application of this technique.
Obtaining the spectrum $y^{[i]}$ (\ref{gevproblem})
and coverage $C^{[i]}$ (\ref{Ccoverage}) (along with cluster localization $D^{[i]}$ from (\ref{wdisclass}))
we actually managed to separate the properties of quantum system itself
and experimental conditions under which the system was tested.
This possibility of separation is the key element
of the approach, because typical ML technique
does not separate them and build a model
combining system properties and experimental conditions together.

\subsection{\label{Distr}$y$--Distribution Estimation. Gauss Quadratures Generalization.}
Considered in previous section idea of
system property and experimental conditions
separation is applicable to a variety of problems.
Consider the simplest one: estimate $y$ distribution
from (\ref{mlproblem}) sample. Evidently, that the pairs
$(y^{[i]},C^{[i]})$
can serve as distribution estimator,  the
states $y^{[i]}$ describe properties of the system itself
and coverages $C^{[i]}$ 
describe  ``experimental conditions'' during $l=[1..M]$ system observations.
\begin{eqnarray}
  P(y^{[i]})&=&C^{[i]} \label{cdist} \\
  \sum\limits_{i=0}^{d_x-1}P(y^{[i]})&=&M
\end{eqnarray}
The (\ref{cdist}) can be considered as Gauss quadratures generalization\cite{nevai,totik}.
If one put to (\ref{mlproblem}) $x^{(l)}_m=Q_m(y^{(l)})$,
then in (\ref{cdist}) the $y^{[i]}$ would be exactly quadrature nodes and $C^{[i]}$
would be quadrature weights.
Regular Gauss $n$--points quadrature
 distribution estimate is using $[0..2n-1]$ moments of $y$  as input,
so its application is limited to distributions with finite $[0..2n-1]$ moments.
The (\ref{cdist}) is much more general in this sense,
it uses some other, $x_q$ random variables moments $<x_qx_r>$ and $<y x_q x_r>$
as input, and only first $y$ moment enter the (\ref{gevproblem}).
This make these new quadratures much better applicable to ML,
because dependent ($y$) and independent ($\mathbf{x}$) variables
are now separated in left-- and right-- hand sides of (\ref{gevproblem}).
In addition to that, because only first $y$ moment enter
the (\ref{gevproblem}) the results can be applied to prediction
of a value out of a non--Gaussian distributions, e.g. the ones with
infinite second $y$ moment.

Another straightforward application of distribution estimation
(\ref{cdist}) may be
information recovery. In Ref. \cite{2015arXiv151101887G}
a basis, obtained as eigenvectors of (\ref{gevproblem}),
was introduced (it is called there not ``Cluster Centers'',
but ``natural basis''),
and image reconstruction was performed in full basis with
all 10000 elements ($d_x=100; d_y=100$) in maximal case.
However, selection as partial basis the states
with maximal coverage (\ref{cdist})
(or cluster relative localization $D^{[i]}/C^{[i]}$ in some cases)
can be a good choice for applications like lossy compression methods
and partial information recovery. This type of application
uses the ``experimental conditions'', the coverage $C^{[i]}$
and cluster localization $D^{[i]}$
as a selection rule for basis components.

\section{\label{numE}Numerical Algorithm}
Numerical instability
similar to the ones we have in Multiple Instance Learning
\cite{2015arXiv151109058G} can also arise here.
The solutions of (\ref{gevproblem}), (\ref{Pi}) and (\ref{ARN})
are invariant with respect to arbitrary linear transform of $\mathbf{x}$
components, but numerical stability of calculations
is drastically different because the condition
number of $G_{qr}$ from (\ref{G})
depend strongly on basis choice\cite{beckermann1996numerical}.
While in \cite{2015arXiv151109058G},
where the moments $<Q_k>$ were used as vector
components of $\mathbf{x}$,
the answer for stable basis choice was rather trivial: for numerical stability
use polynomials $Q_k(x)$ orthogonal with respect
to some measure, e.g. Chebyshev or Legendre polynomials
(see  \cite{polynomialcode}, java implementation of
Chebyshev, Legendre, Laguerre and Hermite
bases
and library description in Appendix A of \cite{2015arXiv151005510G}).
Now, when the $\mathbf{x}$ components can be of different
nature the question of finding linear transform of $\mathbf{x}$,
that give $G_{qr}$ with a
good enough condition number becomes more complicated,
see \cite{beckermann1996numerical} as a good starting point.
However, in this work, we will limit
the number of elements in basis $d_x$ by 20, what
make specific basis choice not that important, compared to say
\cite{2015arXiv151101887G} work, where a problem  with $d_x=d_y=100$,
i.e. 10000 elements in basis have been successfully considered.
Another imortant stability issue is degenerate components
presence in $\mathbf{x}$, when $x_m^{(l)}=x_q^{(l)}$ for $m\ne q$ and all $l$.
Such components make Gramm matrix (\ref{G}) degenerate
and special treatment, like  Tikhonov
regularization\cite{tikhonov1977solutions},
subspace selection by mutual information,
or similar regularization methods may be required.

The algorithm for  $y$ estimators of (\ref{ALS}) or (\ref{ARN})  is this:
Calculate (\ref{G}) and (\ref{yG}) matrices and (\ref{Yq}) vector 
using (\ref{mlproblem}) input data.
Inverse matrix $\left(G\right)_{qr}$ from (\ref{G}),
this matrix is similar to Gramm matrix, but
is build from the components of $\mathbf{x}^{(l)}$ vector.
Finally put all these to (\ref{ALS}) for least squares $y(\mathbf{x})$ estimation
or to (\ref{ARN}) for Radon--Nikodym $y(\mathbf{x})$ estimation.
The (\ref{ALS}) is a linear function of $\mathbf{x}$ components
that posses all the problems typical for least squares --type answers.
The (\ref{ARN}) is a ratio of two quadratic forms
of $\mathbf{x}$ components.
It was shown in Ref. \cite{malha} that in multi--dimensional signal processing
stable estimators can be only of two quadratic forms ratio and
the (\ref{ARN}) is exactly of this form.

If $y$-- distribution is required then solve
generalized eigenvalues problem (\ref{gevproblem}),
obtain $y^{[i]}$ as possible $y$--outcomes, that describe the system itself,
and $C^{[i]}$ from (\ref{Ccoverage}), that describe the testing conditions of the system,
(they both do not depend on vector $\mathbf{x}$ to classify).
Then calculate $\mathbf{x}$--dependent probabilities (\ref{Pi}), these are
squared projection coefficient of a state with given $\mathbf{x}$
to $\psi^{[i]}$ eigenvector.

To show an application of this approach consider a
problem, that can be reduced to Multiple Instance Learning
problem of previous publication
(the case $N=1$ of Multiple Instance Learning
example problem from \cite{2015arXiv151109058G}).
1) For $l=[1..M]$ take random $x$ out of $[-1;1]$ interval.
2) Calculate $y=f(x)$, take this $y$ as $y^{(l)}$.
3) Calculate $x^*=x+R\epsilon$, (where $R$ is a parameter,
$\epsilon$ is uniformly $[-1;1]$ distributed and
$Q_k(x)$ is a polynomial of $k$--th degree),
then take $Q_m(x^*) ; m=[0..d_x-1]$ as the components of input vector $\mathbf{x}^{(l)}$
in (\ref{mlproblem}).
As a function $f$ we take the same three examples from \cite{2015arXiv151109058G}):
\begin{eqnarray}
  f(x)&=&x \label{flin} \\
  f(x)&=&\frac{1}{1+25x^2} \label{frunge} \\
  f(x)&=&\left\{\begin{array}{ll} 0 & x\le 0 \\ 1 & x>0\end{array}\right.
  \label{fstep}
\end{eqnarray}

\begin{figure}[t]
\includegraphics[width=8cm]{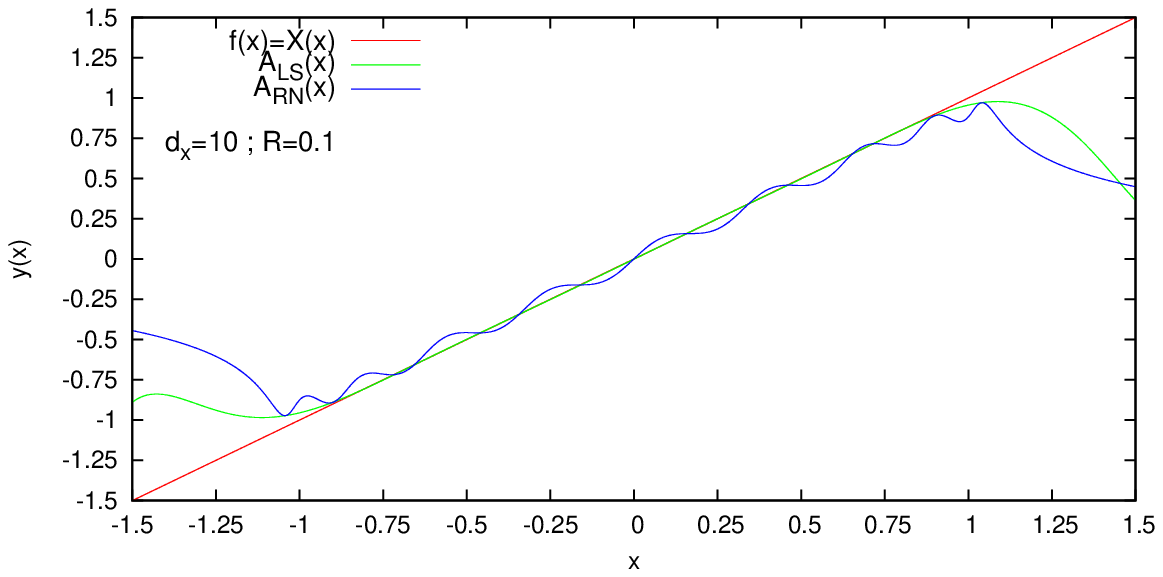}
\includegraphics[width=8cm]{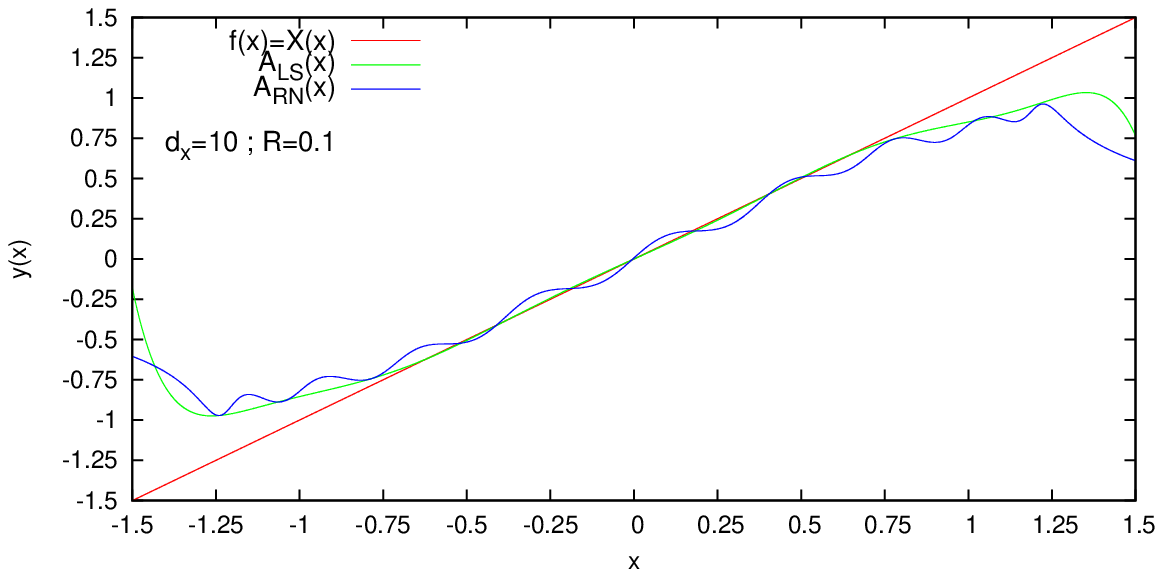}
\includegraphics[width=8cm]{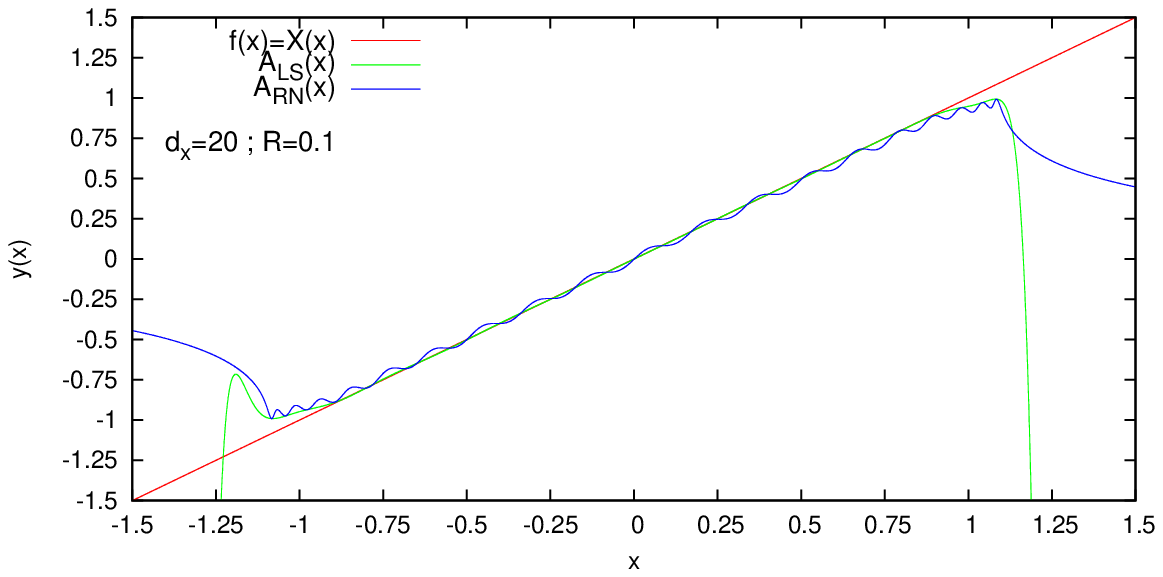}
\includegraphics[width=8cm]{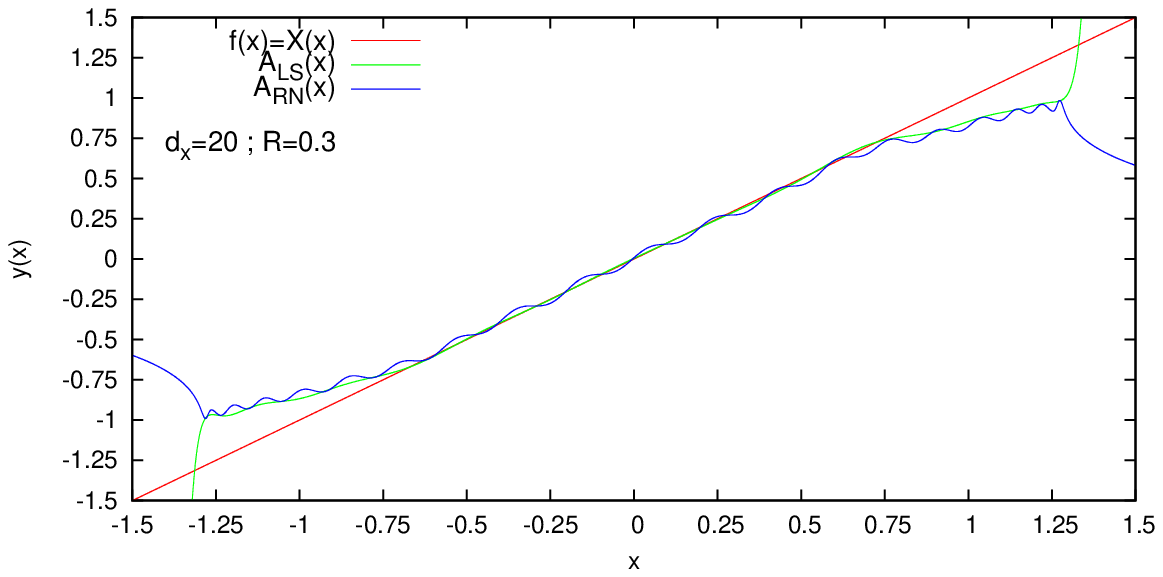}
\caption{\label{fig:flin}
  The $y(x)$ estimation for $f(x)$ from (\ref{flin}).
  }
\end{figure}

\begin{figure}[t]
\includegraphics[width=8cm]{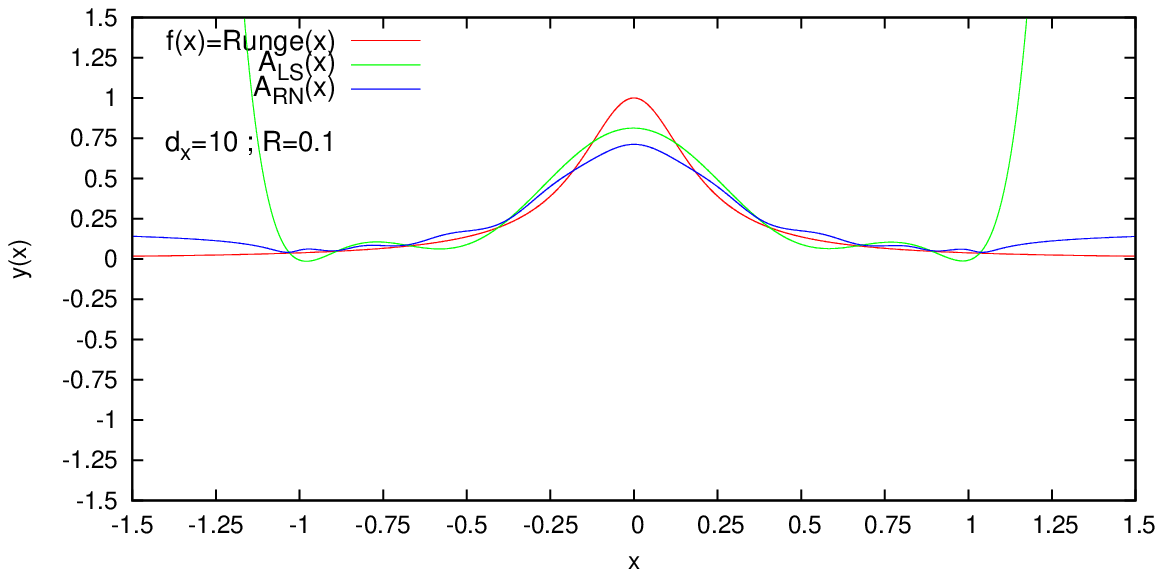}
\includegraphics[width=8cm]{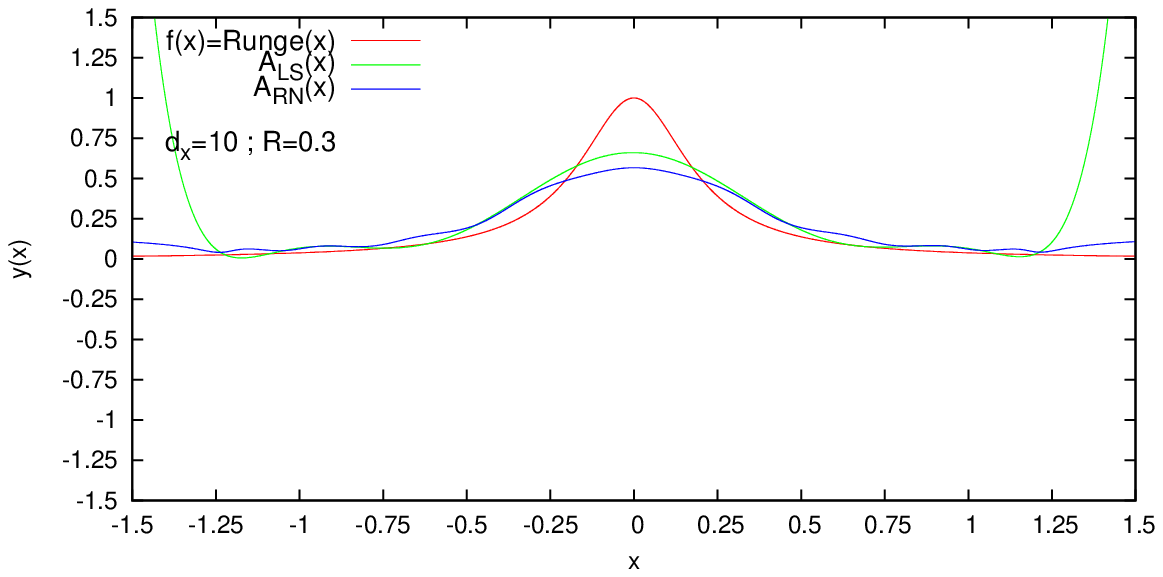}
\includegraphics[width=8cm]{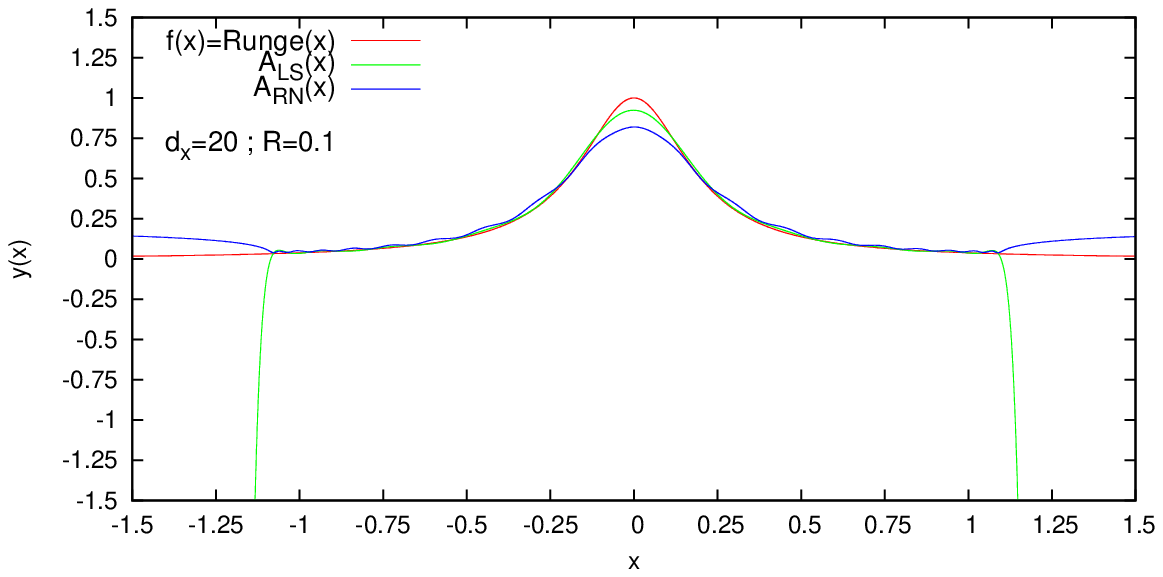}
\includegraphics[width=8cm]{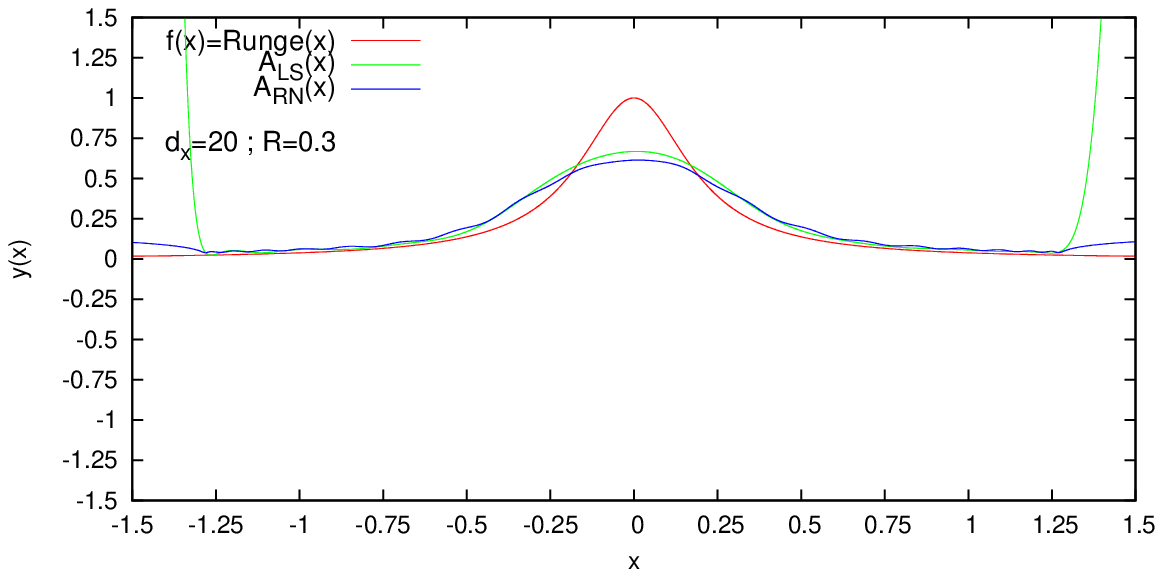}
\caption{\label{fig:frunge}
  The $y(x)$ estimation for $f(x)$ from (\ref{frunge}).
  }
\end{figure}

\begin{figure}[t]
\includegraphics[width=8cm]{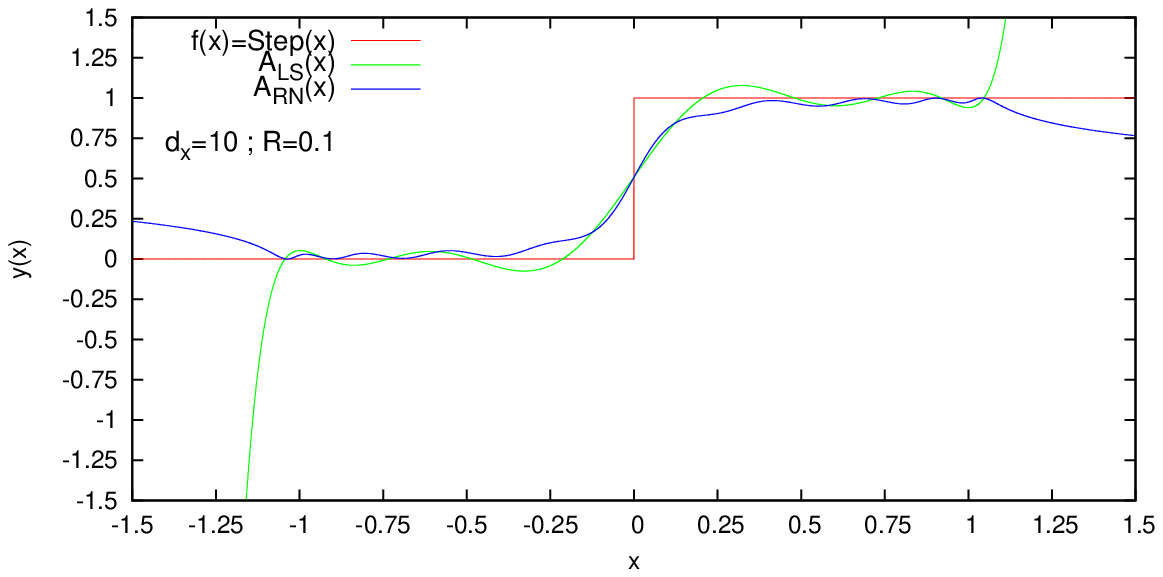}
\includegraphics[width=8cm]{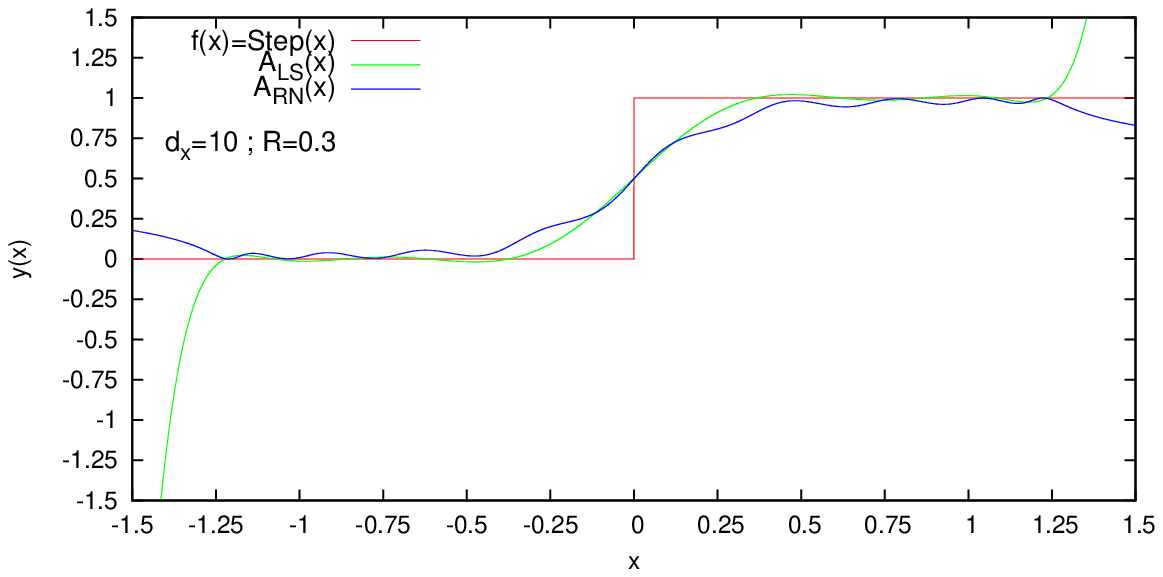}
\includegraphics[width=8cm]{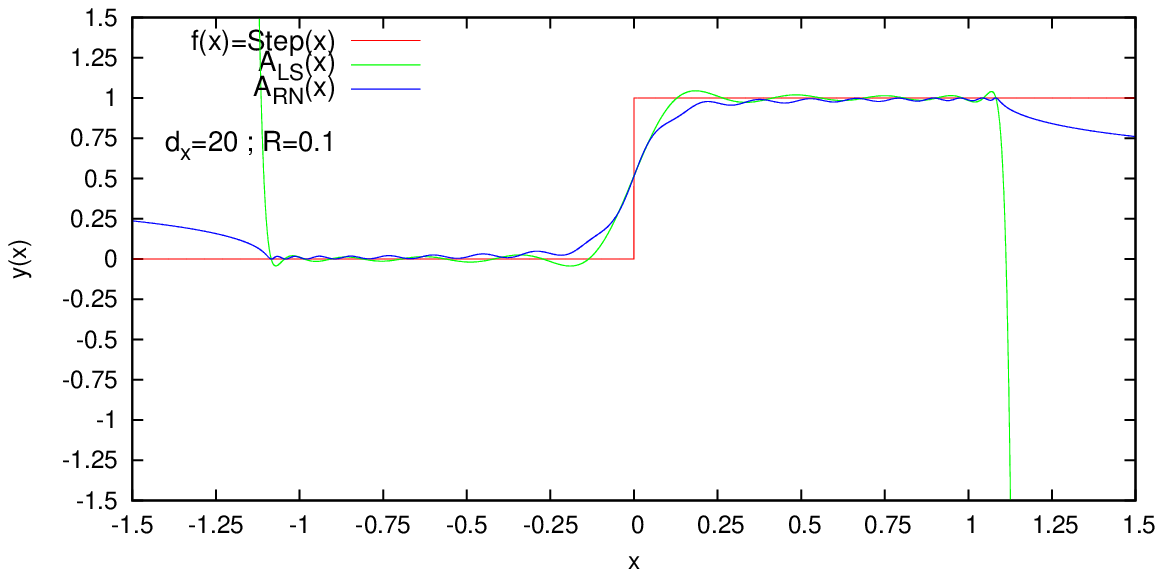}
\includegraphics[width=8cm]{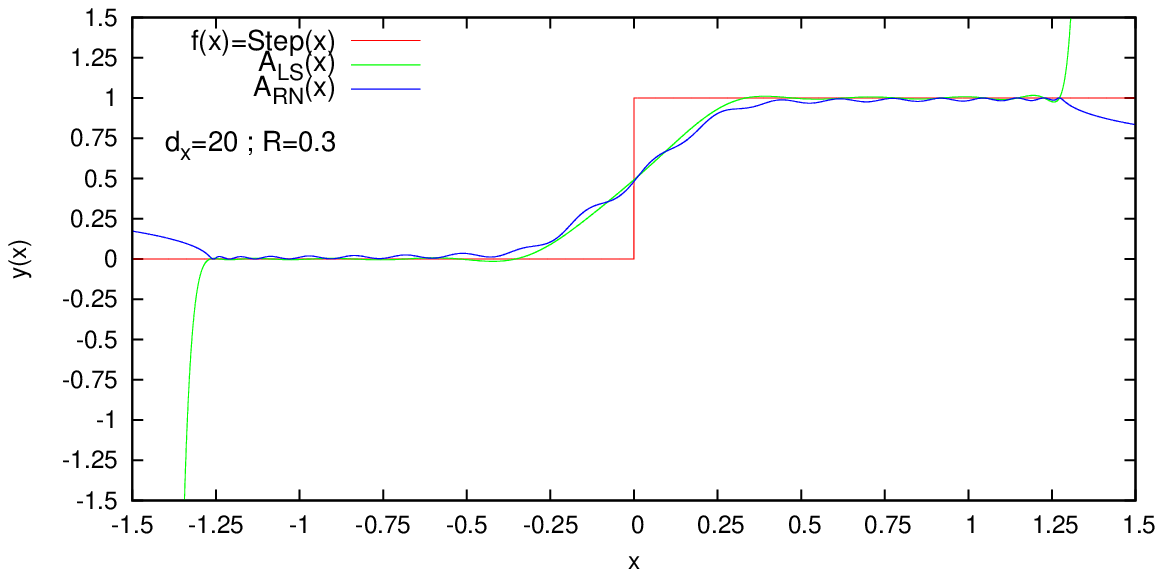}
\caption{\label{fig:fstep}
  The $y(x)$ estimation for $f(x)$ from (\ref{fstep}).
  }
\end{figure}

In Figs. \ref{fig:flin},  \ref{fig:frunge}, \ref{fig:fstep},
the  (\ref{ALS}) and (\ref{ARN}) the answers
are presented for $f(x)$ from (\ref{flin}), (\ref{frunge}) and (\ref{fstep})
respectively 
for $R=\{0.1,0.3\}$
and $d_x=\{10,20\}$. The $x$ range is specially taken slightly
wider that $[-1; 1]$ interval to see possible divergence
outside of measure support.
In most cases Radon--Nikodym answer is superior,
and in addition to that it preserves the sign of $y$.
Least squares approximation is good for special case $f(x)=x$
and typically diverges for $x$ outside of measure support.
The figures are similar to the ones from \cite{2015arXiv151109058G}),
because same data sample was used.

\begin{figure}[t]
\includegraphics[width=14cm]{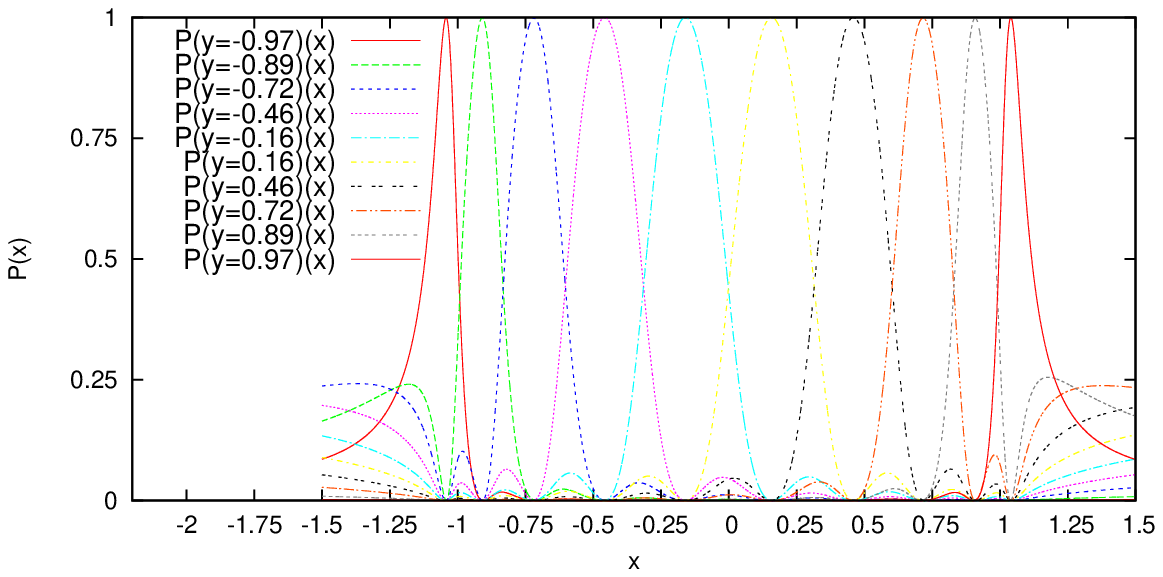}
\caption{\label{fig:Px}
  Probabilities for all $d_x=10$ outcomes of $y^{[i]}$ as a function of $x$ for $f(x)$ from (\ref{flin}).
  }
\end{figure}
An example of numerical estimation of probabilities $P^{[i]}(x)$
is presented in Fig. \ref{fig:Px}
for simplistic case (\ref{flin}).
See  Ref. \cite{polynomialcode}, file
com/polytechnik/ algorithms/ ExampleInterpolationVectorML.scala
for algorithm implementation.

\section{\label{MLdistr}Discussion}
In this work a generalized eigenvectors approach is applied to
ML problem (\ref{mlproblem}).
The interpolatory--type results
of least squares  (\ref{ALS}) and Radon--Nikodym (\ref{ARN}) $y$ value
estimator for a  given $x$ are obtain.
In addition to that distribution estimator of $y$ is obtained:
conditional (\ref{Pi}), at given $\mathbf{x}$,
and unconditional (\ref{cdist}). The last one can be considered
as Gauss quadratures generalization, separating dependent and independent variables
in left-- and right-- hand side of (\ref{gevproblem}).
In this setup ML model can be separated on two parts:
possible $y^{[i]}$ outcomes,
that can be considered as the characteristics of the system itself,
and coverage(probabilities normalized on observations number) $C^{[i]}$, Eq. (\ref{Ccoverage}),
that can be considered as testing conditions of the system the machine is learning from.

Computer code implementing the algorithms is available\cite{polynomialcode}.

\bibliography{LD}

\end{document}